\documentclass[letterpaper, 10 pt, conference]{ieeeconf}  

\IEEEoverridecommandlockouts                              

\usepackage{xcolor}
\usepackage[hidelinks]{hyperref}
\title{\LARGE \bf
Recti-Q: Feature-Space Rectification for Out-of-Distribution-Robust Quantized Perception in Edge Robotics
}

\author{%
\authorblockN{Hamidreza Yaghoubi Araghi$^{*}$, Parastoo Pilevar$^{*}$, and Ming C. Lin}%
\authorblockA{University of Maryland, College Park, MD, USA}%
\noalign{\vskip-2.5em}%
\thanks{$^{*}$Equal contribution. \protect\newline E-mails: \{yaghoubi, pilevar, lin\}@umd.edu. \protect\newline 
Project Page and Code are available at \href{https://hamidrezayaghoubi.github.io/Recti-Q/}
{\protect\textcolor{blue!70!black}{\protect\texttt{Recti-Q}}}}}

\usepackage{algorithm}
\usepackage{algpseudocode}
\usepackage{graphicx}
\usepackage{booktabs}
\usepackage{amsmath}
\usepackage{xspace}
\usepackage{caption}
\usepackage{amssymb}

\newcommand{\eg}{e.g.\xspace}
\begin{document}
\maketitle
\begin{abstract}

Robotic perception pipelines increasingly rely on large vision backbones deployed on SWaP-constrained edge platforms, making post-training quantization (PTQ) attractive for real-time inference. However, while PTQ often preserves clean in-distribution accuracy, we show that it can substantially degrade reliability under deployment-relevant distribution shifts (e.g., sensor noise, severe weather, and novel operating environments), creating a \emph{Quantization-Induced Robustness Gap}. Across foundational vision benchmarks (ImageNet-C and PACS), 4-bit PTQ models exhibit pronounced robustness degradation despite negligible ID accuracy loss. To address this, we propose \textbf{Recti-Q}, a lightweight feature-space rectification framework that freezes the quantized backbone and trains a small classifier-head LoRA adapter using only source data. Recti-Q is architecture-agnostic across CNNs and Transformers, supports efficient teacher-free training, and recovers a significant portion of the lost robustness, in some cases matching or exceeding FP32 performance. At less than 1\% parameter overhead (as small as 6 KB), Recti-Q preserves over 99\% of PTQ memory savings, adds negligible compute, and enables low-bandwidth Over-The-Air (OTA) resilience patching for deployed robotic fleets operating in unpredictable physical environments.

\end{abstract}    
\section{Introduction}
\label{sec:intro}

\begin{figure}[t]
    \centering
    \includegraphics[width=0.8\columnwidth]{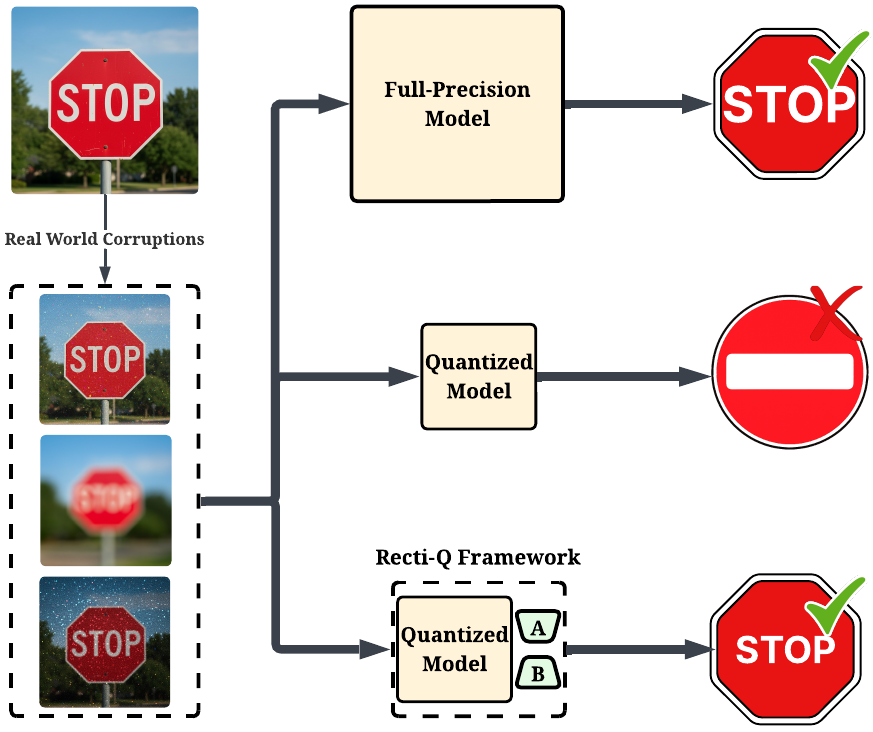}
    \caption{\textbf{Quantization can severely degrade robustness under realistic perturbations.} A clean ``STOP'' sign is corrupted and passed to three visual classification systems representative of an edge robotic perception setting. The full-precision model predicts STOP (correct). The quantized model misreads it as No Entrance (wrong). Our framework, \textbf{Recti-Q}, keeps the quantized backbone and adds a lightweight adapter, restoring the correct STOP prediction while preserving efficiency.}
    \label{fig:overview}
    \vspace*{-1em}
\end{figure}

Visual classification is a core component of many robotic perception pipelines, including autonomous navigation, traffic-sign recognition, and object-centric decision making~\cite{drivingstyles, aiinrobotics, airoboticsrevolutionizing, contetnunderstanding}. These modules are commonly deployed on embedded and edge platforms with tight constraints on memory, latency, and energy~\cite{tanama2023quantized, grainge2024design}, making efficiency a deployment requirement. Meanwhile, robots operate in open environments where inputs are degraded by sensor noise, motion blur, precipitation, illumination changes, and other disturbances. For example, in an autonomous driving stack, a quantized perception module that correctly recognizes a stop sign in clear weather may fail under rain, lens splatter, or glare, and the error can propagate to downstream tracking, prediction, or planning. A perception model must therefore be not only accurate, but also robust to distribution shift. As shown in Figure~\ref{fig:overview}, a full-precision model may handle such perturbations, while its quantized counterpart can fail on the same input.

Post-Training Quantization (PTQ) reduces model size while largely preserving clean accuracy and avoids costly end-to-end retraining, making it attractive for edge deployment~\cite{chen2024comprehensive, grainge2024design, ptqforvit}. However, quantization research typically emphasizes clean accuracy, compression, or throughput, while robustness-preserving alternatives often require full-data QAT or distillation, weakening PTQ's deployment advantage~\cite{chen2024comprehensive}. This motivates a critical question: \emph{do PTQ models remain reliable under distribution shift?} 

\begin{figure}[t]
    \centering
    \includegraphics[width=1.0\columnwidth]{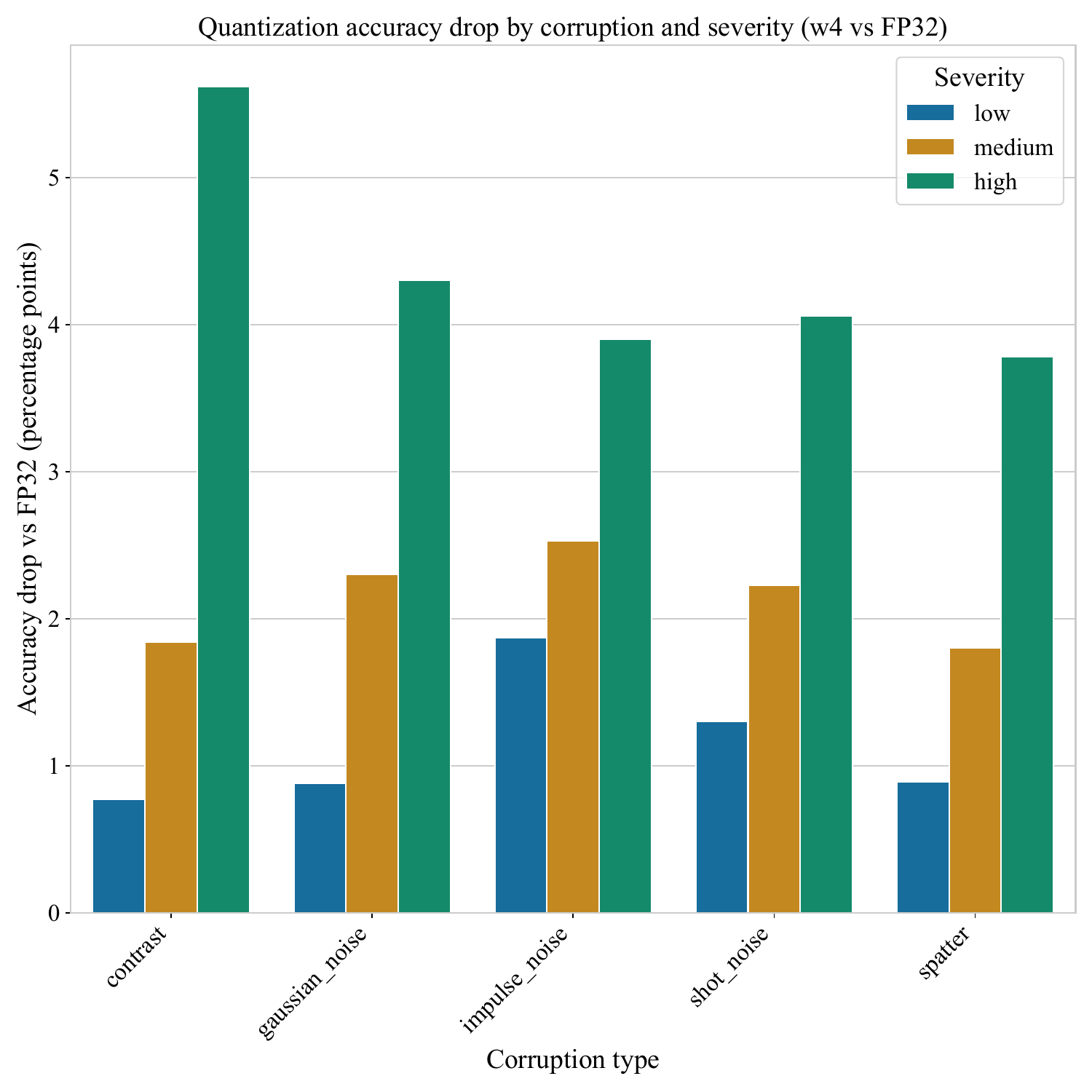}
    \vspace*{-1.5em}
    \caption{\textbf{Quantization degrades robustness under distribution shift.} We plot the accuracy drop of a 4-bit quantized model (relative to its FP32 version) on ImageNet-C at severities 1, 3, and 5 (highest level). The robustness gap grows with corruption severity, showing that PTQ can significantly reduce reliability under realistic input degradation. This issue motivates our \textbf{Recti-Q} framework, which recovers a substantial portion of this lost robustness.}
    \label{fig:imagenet_c_ood_drop}
    \vspace*{-1em}
\end{figure}

In this paper, we identify and quantify this gap, which we term the \emph{Quantization-Induced Robustness Gap}: the loss in out-of-distribution (OOD) robustness caused by PTQ even when clean accuracy remains high. PTQ models can appear reliable on clean inputs (Table~\ref{tab:id_accuracy_ptq}) but, as shown in Figure~\ref{fig:imagenet_c_ood_drop}, degrade sharply once the input distribution shifts. We evaluate on two complementary robustness benchmarks: ImageNet-C~\cite{imagenetc}, which includes realistic corruptions (e.g., noise, blur, contrast change, weather-like effects), and PACS~\cite{pacs}, which probes domain shift across visual styles. Using these foundational benchmarks, we isolate perception-level failure modes and show that Recti-Q addresses the underlying quantization-induced 
feature distortion. 

To repair this gap, we propose \textbf{Recti-Q}, a lightweight framework that decouples compression from robustness. Recti-Q freezes the quantized backbone and adds a tiny, parameter-efficient adapter that rectifies quantization-induced feature perturbations at the classifier head. The adapter is trained only on source data, adds negligible parameter overhead, and supports low-bandwidth Over-The-Air (OTA) robustness patching for deployed robotic fleets by transmitting only the adapter parameters. Recti-Q is architecture-agnostic, working with both CNNs and Transformers, and can be trained with or without a full-precision teacher.

Our results show that Recti-Q recovers a substantial portion of the robustness lost to quantization on both ImageNet-C and PACS, while preserving the memory and efficiency benefits of 4-bit quantization. Our key contributions are:
\begin{itemize}
    \item We identify and quantify the \emph{Quantization-Induced Robustness Gap}, where PTQ models retain strong in-distribution accuracy yet degrade disproportionately under OOD corruption and domain shift.
    \item We introduce \textbf{Recti-Q}, a lightweight and architecture-agnostic framework that repairs a frozen quantized model using a parameter-efficient adapter, enabling low-bandwidth OTA robustness patching.
    \item We show via extensive experiments on ImageNet-C and PACS that Recti-Q substantially improves OOD robustness while retaining the efficiency benefits of 4-bit quantization.
\end{itemize}

\section{Related Work}
\label{sec:related_work}

\subsection{Post-Training Quantization}
Quantization enables deployment in resource-constrained settings by reducing the bit-width of weights and/or activations \cite{chen2024comprehensive}, which is especially important for on-board robotic perception under tight latency, power, and memory budgets. Post-Training Quantization (PTQ) is appealing because it is applied after training, requires little or no fine-tuning data, and avoids full retraining \cite{chen2024comprehensive}. As modern backbones and low-bit regimes become more challenging (\eg, ViTs \cite{dosovitskiy2021imageworth16x16words} and 4-bit quantization), PTQ research has emphasized better calibration \cite{nagel2019datafreequantizationweightequalization, hubara2016quantizedneuralnetworkstraining}, careful parameter choices (e.g., clipping/scales), and data-free/few-shot distillation to preserve clean accuracy \cite{chen2024comprehensive}. In this work, we use a simple, uncalibrated PTQ scheme (\eg, \texttt{torchao} \cite{torchao_arxiv}) as a deployment baseline without a dedicated calibration dataset or weight fine-tuning; as shown by Table~\ref{tab:id_accuracy_ptq}, such PTQ pipelines can appear reliable on clean inputs, motivating closer evaluation under shift.

\subsection{Robustness in Compressed Models}
Efficiency-driven compression has motivated studies of robustness under adversarial perturbations and distribution shift. In embodied systems, quantization-induced feature instability (and, in detection stacks, unstable boxes/tracks) can propagate into downstream subsystems like modular trajectory prediction and motion planning, making robustness beyond clean accuracy a safety requirement. Prior work shows quantization can exacerbate adversarial vulnerability \cite{liu2021zero, goldblum2020adversarially}, and robustness under practical shifts is often evaluated via common corruptions (ImageNet-C \cite{imagenetc}) and domain generalization (\eg, PACS \cite{pacs}). Robustness-preserving quantization methods frequently rely on Robust Quantization-Aware Training (QAT) or robust distillation \cite{chmiel2020robust, goldblum2020adversarially}, which typically require full training data and expensive end-to-end retraining, weakening PTQ's deployment appeal in robotics. Our results in Figure~\ref{fig:imagenet_c_ood_drop} reflect this gap even when clean accuracy is preserved (Table~\ref{tab:id_accuracy_ptq}), motivating a post-training robustness repair on a frozen quantized backbone.

\subsection{Parameter-Efficient Fine-Tuning (PEFT) and Adapters}
Recti-Q builds on Parameter-Efficient Fine-Tuning (PEFT) \cite{PEFTFoundational}, where small auxiliary parameters adapt a frozen backbone. Adapter-based tuning includes bottleneck adapters and low-rank updates \cite{emergentmind_adapterbasedtuning}; LoRA \cite{hu2021lora} is widely used, and PEFT has been applied to domain adaptation and transfer learning \cite{PEFTDomainAdaptation}. Related lightweight strategies include last-layer retraining for group robustness \cite{liu2021justtraintwiceimproving, ghaznavi2023annotation}.

Recent adapter work also targets constraint-induced errors in settings not directly comparable to ours. For example, Quadapter mitigates \emph{GPT-2} quantization issues via activation-outlier handling \cite{park_quadapter_gpt2_quant}, and other work studies adapting \emph{quantized LLMs} \cite{shen_optimal_balance_quant_llm}; these are perplexity-centric and offline, unlike real-time robotic perception under physical-world shift. RepairLLaMA targets \emph{program repair} \cite{silva_repairllama_program_repair}, and Robustness Feature Adapter improves efficiency in \emph{adversarial training} \cite{wu_rfa_efficient_adv_training}, which differs from post-training robustness repair for PTQ vision models under ImageNet-C/PACS shifts.

In contrast, Recti-Q freezes the PTQ vision backbone and localizes adaptation to a small classifier-head module to preserve the efficiency profile and avoid architecture-specific integration. This minimal parameter footprint is well-suited for Over-The-Air (OTA) robustness patching in deployed robotic fleets, a deployment paradigm that remains underexplored in current robot learning literature.
\section{Method}
\label{sec:method}

\begin{figure}[t]
    \centering
    \includegraphics[width=1.0\columnwidth]{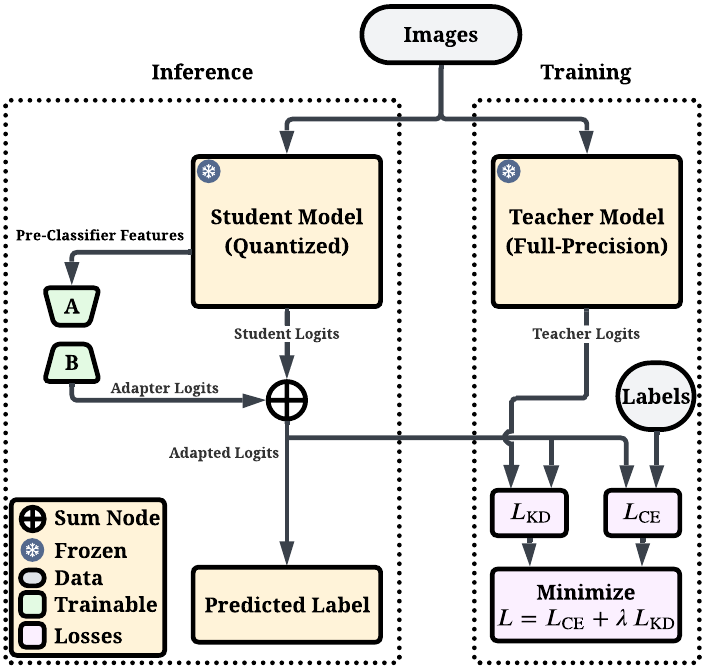}
    \caption{\textbf{Recti-Q training and inference}. A frozen quantized student produces logits that are refined by a trainable LoRA adapter placed at the classifier head. The adapter applies a down-projection $A$ and an up-projection $B$ (scaled by $\alpha/r$) to the pre-classifier features. Training uses only source-domain images and optimizes cross-entropy on labels and a distillation term that compares adapted student logits with teacher logits. The teacher is used only during training. At inference, the quantized backbone and the adapter run, preserving efficiency while restoring {\em robustness}.}
    \label{fig:rectiq_method}
    \vspace*{-1em}
\end{figure}

We start from a pretrained vision model $f$, which we quantize post-training (PTQ) to obtain an efficient, frozen backbone $f_q$. As shown in Table~\ref{tab:id_accuracy_ptq}, PTQ largely preserves standard ID accuracy. However, as we will show in Section~\ref{sec:result}, the same quantized models suffer a substantial drop in robustness on OOD data.

Our goal is to repair this OOD robustness without re-training the full model or incurring high computational costs. We propose Recti-Q, a framework that decouples compression from robustness recovery. The key idea is to freeze the efficient backbone $f_q$ and train a single, parameter-efficient \emph{robustness-repairing adapter}, $g_\phi$, using only source-domain data.

\subsection{Adapter Design: Rectifying Feature Perturbations}
\label{sec:method_1}
We hypothesize that PTQ does not destroy the model's knowledge, but rather perturbs its internal feature representations and decision boundaries. In embodied settings, such unstable representations can translate into inconsistent predictions (and in detection-centric stacks, unstable boxes/tracks) that propagate into downstream autonomy modules. Therefore, simply re-weighting the final, damaged logits is insufficient. Our primary approach is to intervene at the feature level, just before the classifier head.

\textbf{Classifier-LoRA Adapter (Recti-Q):} We introduce a low-rank adapter, $g_\phi$, that operates on the pre-classifier features, $u = \textsc{PreClassifierFeatures}(f_q, x) \in \mathbb{R}^{d}$. This adapter is modeled as a standard LoRA module \cite{hu2021lora}, consisting of a down-projection $A \in \mathbb{R}^{r \times d}$ and an up-projection $B \in \mathbb{R}^{C \times r}$ (with rank $r$ and $C$ classes):
\begin{equation}
    g_\phi(u) = B(A(u)) \cdot (\alpha/r),
\end{equation}
so $g_\phi(u) \in \mathbb{R}^{C}$ is a logit-space correction computed from pre-classifier features. The final logits are
\begin{equation}
z = z_q + g_\phi(u) 
= f_q(x) + g_\phi(u).
\end{equation}
Equivalently, if the quantized head is $W_q \in \mathbb{R}^{C \times d}$ with $z_q = W_q u$, Recti-Q induces a low-rank update to the head, $z = (W_q + (\alpha/r)BA)\,u$, while keeping $W_q$ frozen. Conceptually, this low-rank update acts as a rapid feature-space calibration, realigning representations that become distorted when the quantized weights process out-of-domain physical sensor inputs (e.g., camera glare or weather artifacts). Crucially, the up-projection $B$ is zero-initialized, so at the start of training, $z = z_q$. This ensures a stable training process, where the adapter learns to gently correct the quantized model's predictions.

\begin{table}[t!]
\centering
\small 
\setlength{\tabcolsep}{5pt} 
\caption{
    \textbf{PTQ preserves ID accuracy.}
    We evaluate top-1 accuracy of FP32 models and their W4 quantized counterparts on standard ID test splits. On both PACS and ImageNet, the W4 models show negligible degradation. This establishes that standard ID metrics \textit{alone} are insufficient to evaluate \textbf{model robustness}, as they fail to capture the substantial OOD degradation, to be shown in Section~\ref{sec:result}.
}
\label{tab:id_accuracy_ptq}
\begin{tabular}{l c c c c}
\toprule
& \multicolumn{2}{c}{PACS (ID) (\%)} & \multicolumn{2}{c}{ImageNet (ID) (\%)} \\
\cmidrule(lr){2-3} \cmidrule(lr){4-5}
Model & FP32 Acc. & W4 Acc. & FP32 Acc. & W4 Acc. \\
\midrule
ResNet50 & 94.99 & 94.87 & 80.38 & 80.31 \\
DeiT-t & 94.48 & 94.26 & 72.16 & 71.54 \\
DeiT-s & 96.38 & 96.38 & 79.85 & 78.94 \\
DeiT-b & 97.21 & 97.55 & 81.98 & 81.50 \\
\bottomrule
\end{tabular}
\end{table}

\subsection{Recti-Q Training and Inference}
The complete Recti-Q framework is depicted in Figure~\ref{fig:rectiq_method}. During training, the quantized student $f_q$ is frozen. The full-precision teacher $f_t$ is also kept frozen and is only required if knowledge distillation is applied ($\lambda > 0$). Our framework fully supports a ``teacher-free'' mode ($\lambda=0$) for maximum training efficiency. In all cases, only the lightweight adapter $g_\phi$ is trained. The full training procedure is detailed in Algorithm~\ref{alg:rectiq-train}.

\begin{algorithm}[t]
\caption{\textbf{Recti-Q} training on a frozen quantized backbone with a classifier-head LoRA adapter and optional KD. In teacher-free runs, we set $\lambda=0$ and skip computing $s_t$ and $L_{\text{KD}}$ in implementation.}
\label{alg:rectiq-train}
\begin{algorithmic}[1]
\Require Frozen quantized model $f_q$; optional frozen teacher $f_t$; trainable LoRA adapter $g_\phi$; source data $\mathcal{D}_{\text{src}}$; validation data $\mathcal{D}_{\text{val}}$; distillation weight $\lambda$; temperature $T$
\For{epoch $=1,\dots,E$}
  \For{minibatch $(x,y)$ in $\mathcal{D}_{\text{src}}$}
    \State $z_q \gets f_q(x)$
    \State $u \gets \textsc{PreClassifierFeatures}(f_q,x)$ \Comment{$u$ is cached from the same forward pass}
    \State $z \gets z_q + g_\phi(u)$
    \State $L_{\text{CE}} \gets \mathrm{CE}(z,y)$
    \State $s_z \gets \mathrm{softmax}(z/T)$
    \State $s_t \gets \mathrm{softmax}(f_t(x)/T)$
    \State $L_{\text{KD}} \gets \mathrm{KL}(s_t \,\|\, s_z)\,T^2$ 
    \State $L \gets L_{\text{CE}} + \lambda L_{\text{KD}}$
    \State Update $\phi$ with AdamW on $L$
  \EndFor
  \State Validate on $\mathcal{D}_{\text{val}}$; keep the best $\phi$
\EndFor
\State \Return trained adapter $g_\phi$
\end{algorithmic}
\end{algorithm}

At inference, the teacher model is discarded. The forward pass consists of a single pass through the frozen quantized backbone $f_q$ and the tiny, trained adapter $g_\phi$. This process, detailed in Algorithm~\ref{alg:rectiq_infer}, adds negligible parameter overhead while restoring OOD robustness.

\begin{table*}[t]
\centering
\small
\setlength{\tabcolsep}{5pt} 
\caption{
    \textbf{Recti-Q Recovers OOD Robustness (PACS) While Preserving Efficiency.}
    Our method (Ours) not only recovers the accuracy of the degraded W4 baseline but in some cases \textbf{exceeds the original FP32 performance} (e.g., DeiT-t on cartoon, 73.34\% vs 73.08\%). 
    Crucially, the final three columns show the complete efficiency story: our total model size (`Ours`) \textbf{retains over 99\% of the W4 model's memory savings}, demonstrating a clear, practical win for both \textbf{robustness} and \textbf{efficiency}.
}
\label{tab:main_results_pacs}
\begin{tabular}{l l | c c c | c c | c c c}
\toprule
& & \multicolumn{3}{c}{\textbf{OOD Accuracy (\%)}} & \multicolumn{2}{c}{\textbf{Robustness Delta (pp)}} & \multicolumn{3}{c}{\textbf{Model Size (MB)}} \\
\cmidrule(lr){3-5} \cmidrule(lr){6-7} \cmidrule(lr){8-10}
Model & Domain & FP32 & W4 & \textbf{Ours} & W4 vs FP32 (Gap) & Ours vs W4 (Recov.) & FP32 & W4 & \textbf{Ours} \\
\midrule
DeiT-s & sketch             & 68.59 & 64.57 & 65.74 & -4.02 & \textbf{+1.17} & 82.71 & 25.64 & \textbf{25.74} \\
DeiT-t & cartoon            & 73.08 & 71.72 & \textbf{73.34} & -1.36 & \textbf{+1.62} & 21.13 & 15.59 & \textbf{15.64} \\
DeiT-t & art\_painting      & 74.90 & 73.63 & \textbf{75.15} & -1.27 & \textbf{+1.52} & 21.13 & 15.59 & \textbf{15.64} \\
DeiT-s & art\_painting      & 88.38 & 87.21 & 88.28 & -1.17 & \textbf{+1.07} & 82.71 & 25.64 & \textbf{25.74} \\
ResNet50 & sketch           & 72.46 & 72.42 & \textbf{73.30} & -0.04 & \textbf{+0.88} & 90.03 & 90.03 & 90.53 \\  
\bottomrule
\end{tabular}   
\end{table*}

\begin{table*}[t]
\centering
\small
\setlength{\tabcolsep}{5pt} 
\caption{
    \textbf{Recti-Q Recovers OOD Robustness on ImageNet-C While Preserving Efficiency}. Our method (Ours) is applied to the W4 baseline on the most challenging corruptions from ImageNet-C (severity 5). The table demonstrates a strong and consistent recovery, \textbf{closing nearly 50\% of the robustness gap} on the worst-case contrast corruption (recovering +2.68pp of a -5.62pp drop). Crucially, the final three columns show this \textbf{robustness} gain is achieved with minimal cost: the total model size ('Ours') \textbf{retains over 99\% of the W4 model's memory savings}, validating our framework's practical efficiency.
}
\label{tab:main_results_imagenetc}
\begin{tabular}{l l | c c c | c c | c c c}
\toprule
& & \multicolumn{3}{c}{\textbf{OOD Accuracy (\%)}} & \multicolumn{2}{c}{\textbf{Robustness Delta (pp)}} & \multicolumn{3}{c}{\textbf{Model Size (MB)}} \\
\cmidrule(lr){3-5} \cmidrule(lr){6-7} \cmidrule(lr){8-10}
Model & Corruptions & FP32 & W4 & \textbf{Ours} & W4 vs FP32 (Gap) & Ours vs W4 (Recov.) & FP32 & W4 & \textbf{Ours} \\
\midrule
DeiT-s & contrast           & 39.46 & 33.84 & 36.52 & -5.62 & \textbf{+2.68} & 84.17 & 26.16 & \textbf{26.50} \\
DeiT-s & gaussian\_noise    & 33.18 & 28.88 & 29.30 & -4.30 & \textbf{+0.42} & 84.17 & 26.16 & \textbf{26.50} \\
DeiT-s & shot\_noise        & 30.36 & 26.30 & 26.67 & -4.06 & \textbf{+0.37} & 84.17 & 26.16 & \textbf{26.50} \\
DeiT-s & impulse\_noise     & 32.82 & 28.92 & 29.19 & -3.90 & \textbf{+0.27} & 84.17 & 26.16 & \textbf{26.50} \\
ResNet50 & impulse\_noise   & 19.83 & 19.68 & \textbf{27.04} & -0.15 & \textbf{+7.37} & 97.79 & 91.02 & \textbf{91.77} \\ 
\bottomrule
\end{tabular}
\vspace*{-0.5em}
\end{table*}

\begin{algorithm}[t]
\caption{Inference with \textbf{Recti-Q}}
\label{alg:rectiq_infer}
\begin{algorithmic}[1]
\Require Frozen $f_q$, trained $g_\phi$, input $x$
\State $z_q \gets f_q(x)$
\State $u \gets \textsc{PreClassifierFeatures}(f_q,x)$ \Comment{$u$ is cached from the same forward pass}
\State $z \gets z_q + g_\phi(u)$
\State \textbf{return} $\hat{y}=\arg\max z$
\end{algorithmic}
\end{algorithm}

\subsection{Optimization Objective}
Only adapter parameters are updated. The quantized backbone and, when used, the float teacher remain frozen and in evaluation mode. Training uses standard classification loss on allowed source data. When a float teacher is available, we add a temperature-scaled distillation term to preserve class structure learned by the original model~\cite{hinton2015distilling}.

Plainly, the objective is defined as
\begin{equation}
    L = L_{\mathrm{CE}} + \lambda \, L_{\mathrm{KD}},
\end{equation}
where $L_{\mathrm{CE}}$ is the cross-entropy loss on the source labels, 
$L_{\mathrm{KD}}$ is the Kullback--Leibler divergence between the adapted and teacher predictions 
(with temperature $T$), and $\lambda$ controls the strength of the distillation. 
In teacher-free runs, we set $\lambda = 0$. 
This provides a practical and highly efficient adaptation scenario that avoids the computational cost and memory overhead of loading and running the full-precision teacher model. We analyze the performance of this teacher-free approach in our experiments (see Section~\ref{sec:result}), demonstrating its efficacy as a minimal-cost solution.

We use the AdamW optimizer~\cite{adamw} with a small learning rate, light weight decay, and a short cosine schedule (converging in just a few epochs, typically requiring only minutes of numerical optimization on a standard consumer GPU or local edge server, facilitating rapid depot-level adaptation), with the same preprocessing and augmentations as in the source training recipe.

\subsection{Source-only Adaptation Protocol}
We follow leakage-free, community-standard evaluation. For ImageNet-C, adapters are trained only on a fixed 5\% class-balanced subsample of ImageNet-1k train and validated on ImageNet-1k val; evaluation is on ImageNet-C across corruption types and severities~\cite{imagenetc}. For PACS, we use leave-one-domain-out: for each target domain, adapters are trained on the other three source domains, validated on source splits, and evaluated once on the held-out target~\cite{pacs}. No target images are used for training or model selection. This source-only constraint is practical and challenging, as it forgoes any target-domain data, a paradigm explored in other robustness frameworks \cite{saberi2023out}. In embodied AI, this protocol reflects ``Lab-to-Wild'' or ``Sim-to-Real'' deployment, where a robot must operate in unseen environments without prior target-domain calibration.

\subsection{Complexity and Deployment}
Recti-Q concentrates adaptation at the head, so the added parameter overhead is small relative to the backbone. For a classifier with feature dimension $d$ and $C$ classes, the LoRA path adds roughly $r(d + C)$ weights for rank $r$, much smaller than the full head. The backbone remains quantized and unchanged, so the total memory footprint is preserved while robustness is improved. This also enables low-bandwidth Over-The-Air (OTA) robustness patching for deployed robotic fleets: once a base quantized model is on-device, only the small adapter parameters need to be transmitted to update the system for new environments. Because the adapter operates solely on the low-dimensional pre-classifier features, it requires only a trivial number of multiply-accumulate (MAC) operations ($r \times d$ and $C \times r$). 
Compared with the billions of MACs executed by the visual backbone, this additional computation is negligible. 
\section{Results}
\label{sec:result}

\subsection{Experimental Setup}

\noindent\textbf{Datasets and Protocols.}
Our experiments are conducted on two standard out-of-distribution (OOD) benchmarks, following strict source-only evaluation protocols.

\begin{itemize}
    \item \textbf{ImageNet-C:} For corruption robustness, we use ImageNet-C~\cite{imagenetc}. We follow a resource-efficient, ``leakage-free'' protocol: adapters are trained on a fixed 5\% class-balanced subsample of the ImageNet-1k \textit{train} set and validated on the standard ImageNet-1k \textit{val} split. All final results are reported on the ImageNet-C test set. No corrupted data is seen during training or validation.
    
    \item \textbf{PACS:} For domain generalization, we use the PACS dataset~\cite{pacs}, which contains four distinct domains (Photo, Art, Cartoon, Sketch). We follow the standard leave-one-domain-out (LODO) protocol. For each domain held out as the target, we train adapters using only the data from the remaining three source domains.
\end{itemize}


\noindent\textbf{Models and Baselines.}
We evaluate on both CNNs and Transformers to show broad applicability: ResNet-50, DeiT-Tiny (DeiT-t), DeiT-Small (DeiT-s), and DeiT-Base (DeiT-b). All models use pretrained weights from \texttt{timm} \cite{wightman2019timm}. We compare three primary configurations:

\begin{itemize}

    \item \textbf{FP32 (Reference):} The original, full-precision model, which serves as our ``gold standard'' for robustness. 

    \item \textbf{PTQ (W4) (Problem Baseline):} The frozen backbone after applying weight-only 4-bit quantization using \texttt{torchao}. We use the \texttt{Int4WeightOnlyConfig} with \texttt{use\_hqq=True}. This enables Half-Quadratic Quantization (HQQ), a simple, optimization-based technique that does not require a calibration dataset. This is a crucial design choice: by using a strong, calibration-free method, we establish a rigorous baseline and avoid introducing data-driven artifacts that could confound the OOD robustness evaluation. This model therefore represents a ``pure'' PTQ baseline and highlights the true ``Quantization-Induced Robustness Gap.'' We focus our main analysis on W4 as a strict stress test, since it exhibits the most severe robustness degradation; repairing this regime validates Recti-Q under aggressive compression. Furthermore, W4 simulates the extreme compression regimes increasingly demanded by memory-bandwidth-bound robotic edge accelerators (e.g., edge TPUs or embedded GPUs) deployed in SWaP-constrained autonomous systems.
    
    \item \textbf{Recti-Q (Ours):} The identical W4-quantized backbone augmented with our lightweight adapter. The backbone remains frozen, and only the adapter parameters are trained.
\end{itemize}

\noindent\textbf{Implementation Details.}
Unless stated otherwise, all Recti-Q results use our Classifier-LoRA Adapter with a rank $r=64$ and $\alpha=16$ (applied with $\alpha/r$ scaling). We train all adapters using the AdamW optimizer~\cite{adamw} with a learning rate of $3 \times 10^{-4}$ and weight decay of $1 \times 10^{-4}$. Adapters for ImageNet-C are trained for 5 epochs; adapters for PACS are trained for 5-10 epochs. We use a training batch size of 128 and a validation/test batch size of 256. For Knowledge Distillation (KD), we use the FP32 model as the teacher, a temperature $T=4$, and a distillation weight $\lambda \in \{0, 0.5, 1.0\}$. The $\lambda=0$ case represents our ``teacher-free'' variant, which relies solely on the cross-entropy loss.

\subsection{The Quantization-Induced Robustness Gap}
\label{sec:quantization_gap}

We begin our experimental analysis by investigating the effect of 4-bit Post-Training Quantization (W4 PTQ) on model robustness.

\noindent\textbf{The ID Accuracy is Preserved, but Misleading.} First, we establish a critical, but deceptive, baseline. We evaluate our models on standard ID test sets. As shown in Table \ref{tab:id_accuracy_ptq}, applying W4 PTQ results in a negligible accuracy change. On the ImageNet validation set, the accuracy drop is less than 
1.0 
percentage points for all DeiT models. On the PACS ID split, performance is nearly identical, with DeiT-b even showing a minor improvement. A developer looking only at these standard metrics would conclude the quantization was successful and robust to deploy.

\noindent\textbf{The ``Robustness Gap'': OOD Robustness Degrades.} However, this perceived robustness shatters when the models are evaluated under out-of-distribution (OOD) shifts. This disparity is evident in two distinct OOD settings. For brevity, we report representative worst-case domains and corruptions here to highlight the phenomenon.

\begin{itemize} 
    \item \textbf{On Common Corruptions (ImageNet-C):} We first analyze performance on the ImageNet-C benchmark. Table \ref{tab:imagenet_c_ood_drop} quantifies the accuracy drop for the worst-case corruptions at severity 5. For the DeiT-s model, this drop is severe, reaching -5.62 percentage points on the `contrast` corruption. 
    In an embodied context, such a substantial drop may increase the risk of perception failures under low-light or foggy conditions, even when clean-input performance appears reliable. 

    \item \textbf{On Domain Shift (PACS):} This robustness loss is not limited to corruptions. We observe a similar pronounced drop in the PACS domain-shift benchmark. Table \ref{tab:pacs_ood_drop} quantifies this degradation on the four domains that exhibited the largest accuracy drop. On the DeiT-s sketch domain, W4 quantization causes the model to lose over 4 percentage points of accuracy, erasing a significant portion of its generalization capability.
\end{itemize}

Taken together, Table \ref{tab:imagenet_c_ood_drop} and Table \ref{tab:pacs_ood_drop} provide clear evidence of a Quantization-Induced Robustness Gap. While PTQ successfully preserves ID accuracy, it creates a severe and previously under-explored vulnerability to OOD shifts. This gap motivates the central goal of our work: to repair this robustness degradation.

\begin{table}[t!] 
\centering 
\small 
\setlength{\tabcolsep}{5pt}
\caption{
    \textbf{PTQ substantially degrades OOD robustness on the PACS domain-shift benchmark.} 
    We show the accuracy of the FP32 ``gold standard'' vs. the W4 PTQ baseline on held-out OOD domains. The table highlights the worst-case ``Model-Domain'' pairs, showing a significant drop in robustness (e.g., -4.02 pp for DeiT-s on sketch) that was not visible in the ID data (Table \ref{tab:id_accuracy_ptq}).
}
\label{tab:pacs_ood_drop} 
\begin{tabular}{l l c c c} 
\toprule 
& & \multicolumn{3}{c}{\textbf{OOD Accuracy Analysis}} \\ 
\cmidrule(lr){3-5}
Model & Domain & FP32 (\%) & W4 (\%)  & W4 Drop (pp) \\ 
\midrule 
DeiT-s & sketch & 68.59 & 64.57 & \textbf{-4.02} \\ 
DeiT-t & cartoon & 73.08 & 71.72 & \textbf{-1.36} \\ 
DeiT-t & art\_painting & 74.90 & 73.63 & \textbf{-1.27} \\ 
DeiT-s & art\_painting & 88.38 & 87.21 & \textbf{-1.17} \\ 
\bottomrule 
\end{tabular} 
\end{table}

\begin{table}[t!] 
\centering 
\small 
\setlength{\tabcolsep}{4pt}
\caption{
    \textbf{PTQ substantially degrades OOD robustness on ImageNet-C.}
    This table highlights the worst-case drops for a selection of corruptions at severity 5 (for DeiT-s). This severe drop was not visible in the ID data (Table \ref{tab:id_accuracy_ptq}).
}
\label{tab:imagenet_c_ood_drop} 
\begin{tabular}{l l c c c} 
\toprule 
& & \multicolumn{3}{c}{\textbf{OOD Accuracy Analysis}} \\ 
\cmidrule(lr){3-5}
Model & Corruptions & FP32 (\%) & W4 (\%) & W4 Drop (pp) \\ 
\midrule 
DeiT-s & contrast & 39.46 & 33.84 & \textbf{-5.62} \\ 
DeiT-s & gaussian\_noise & 33.18 & 28.88 & \textbf{-4.30} \\ 
DeiT-s & shot\_noise & 30.36 & 26.30 & \textbf{-4.06} \\ 
DeiT-s & impulse\_noise & 32.82 & 28.92 & \textbf{-3.90} \\ 
\bottomrule 
\end{tabular} 
\end{table}

\noindent\textbf{Breakdown point.}
On DeiT-s, W3 retains 73.61\% ID accuracy, and Recti-Q raises mean OOD accuracy across the four Table~\ref{tab:main_results_imagenetc} corruptions from 12.05\% to 14.24\% (+2.19 pp). At W2, ID accuracy collapses to 4.61\% and recovery falls to +0.12 pp (0.21\% to 0.33\%), identifying W2 as the empirical breakdown point between correctable feature distortion and information loss. 

\subsection{Recti-Q Recovers OOD Robustness}

Having established the ``Quantization-Induced Robustness Gap'' in Section \ref{sec:quantization_gap}, we now demonstrate our method's ability to repair it. We present our main results for both domain shift (PACS) in Table \ref{tab:main_results_pacs} and common corruptions (ImageNet-C) in Table \ref{tab:main_results_imagenetc}.

\noindent\textbf{Performance on Domain Shift (PACS).} The results on the PACS benchmark are compelling. As shown in Table \ref{tab:main_results_pacs}, our Recti-Q method not only recovers the accuracy drop but, in several cases, exceeds the original FP32 ``gold standard'' performance. For instance, on the cartoon and art\_painting domains, the DeiT-t model, when augmented with our adapter, achieves 73.34\% and 75.15\% accuracy, surpassing the FP32 performance of 73.08\% and 74.90\%, respectively. 
These gains are best interpreted as calibration of quantization-induced shifts in feature geometry and the decision boundary: Recti-Q restores robustness supported by the frozen backbone rather than acquiring semantic understanding of unseen classes. 
Crucially, the final three columns quantify our core Recti-Q claim. The full FP32 DeiT-s model (82.71 MB) is compressed to just 25.64 MB by W4 PTQ. Our method adds only 0.10 MB (25.74 MB total), retaining over 99.6\% of the memory savings while restoring, and even improving robustness.

\noindent\textbf{Performance on Common Corruptions (ImageNet-C).} We observe a similarly strong success on the challenging ImageNet-C benchmark, shown in Table \ref{tab:main_results_imagenetc}. Here, we focus on the most severe corruptions (severity=5) for the DeiT-s model.

The recovery is both significant and consistent. For the contrast corruption, which causes the largest drop of -5.62 pp, our method recovers +2.68 pp, closing nearly 50\% of the robustness gap. While the recovery magnitude varies by corruption type, Recti-Q provides consistent, positive gains across all other tested corruptions, demonstrating a robust repair mechanism.

This robustness is again achieved with near-zero cost. The final three columns show the total model size (26.50 MB) remains almost identical to the W4 baseline (26.16 MB). Taken together, these results validate our central hypothesis: by decoupling compression and robustness, we can use a tiny, parameter-efficient adapter to repair the OOD robustness of a frozen quantized model, all while retaining over 99\% of the memory savings as discussed in Section~\ref{sec:method}.

\subsection{Ablation Studies}

We conduct ablation studies to validate our key design choices: the necessity of the FP32 teacher (KD loss), the superiority of our feature-space adapter, and the effect of adapter rank.

\subsubsection{Is the FP32 Teacher Necessary?}

A key component of our efficiency claim is the ability to train without a large, full-precision teacher model. We test this by comparing our full method (using KD with $\lambda > 0$) against a ``teacher-free'' variant that trains only with cross-entropy ($\lambda = 0$).
We focus our analysis on the four worst-case domains from our main results. The results, shown in Table \ref{tab:ablation_kd}, show that for the DeiT-t model, the teacher-free ($\lambda=0$) variant is the strongest performer, achieving the highest accuracy on both the art\_painting (+1.52 pp gain) and cartoon (+1.62 pp gain) domains. For DeiT-s, the teacher provides a clear benefit on the sketch domain (+1.17 pp vs. +0.08 pp), but the teacher-free model is still highly effective on art\_painting (recovering +1.07 pp).

This is an important finding: while the FP32 teacher can provide a boost in specific cases, our Recti-Q framework is highly effective even in its most efficient, teacher-free configuration. This confirms that our method is practical for scenarios where a teacher model is unavailable or too costly to run.

\begin{table}[t!]
\centering
\small
\setlength{\tabcolsep}{4pt} 
\caption{
    \textbf{Ablation: The FP32 teacher (KD) is not always required.}
    We compare the teacher-free (\textbf{$\lambda=0$}) Recti-Q with two teacher-guided variants ($\lambda=0.5$, $\lambda=1.0$) on the worst-case PACS domains. The teacher-free model is highly competitive and even achieves the \textbf{best performance} for DeiT-t.}
\label{tab:ablation_kd}
\begin{tabular}{l l | c | c c c}
\toprule
& & \textbf{Baseline} & \multicolumn{3}{c}{\textbf{Ours (Accuracy \%)}} \\
\cmidrule(lr){3-3} \cmidrule(lr){4-6}
Model & Domain & W4 Acc. & \textbf{$\lambda=0$} & $\lambda=0.5$ & $\lambda=1.0$ \\
\midrule
DeiT-s & sketch & 64.57 & 64.65 & 65.49 & \textbf{65.74} \\
DeiT-s & art\_painting & 87.21 & \textbf{88.28} & 87.68 & 87.71 \\
\midrule
DeiT-t & cartoon & 71.72 & \textbf{73.34} & 72.87 & 72.85 \\
DeiT-t & art\_painting & 73.63 & \textbf{75.15} & 74.41 & 74.24 \\
\bottomrule
\end{tabular}
\end{table}

\subsubsection{Is Feature-Space Rectification Necessary?}

Our core hypothesis is that we must ``rectify feature perturbations'' (Section~\ref{sec:method_1}). To prove this, we compare our Classifier-LoRA Adapter (which operates on features) against a Logit-Space Adapter (a simple MLP baseline that operates on the final, damaged logits).

As shown in Table \ref{tab:ablation_adapter_type}, the baseline provides no measurable gain. On the DeiT-s sketch domain, the Logit-Space Adapter provides zero recovery (+0.00 pp) compared to the +1.17 pp gain from our feature-space method. This confirms our hypothesis: the OOD degradation is a deep feature-level problem, and simply re-weighting the final logits is an ineffective solution.

\begin{table}[h!]
\centering
\small
\caption{
    \textbf{Ablation: Feature-Space (Ours) vs. Logit-Space (Baseline).}
    We compare our Recti-Q (Feature-LoRA) against a baseline MLP adapter on the final logits. The results, on the worst-case DeiT-s sketch domain, prove our feature-space rectification hypothesis is necessary.
}
\label{tab:ablation_adapter_type}
\begin{tabular}{l c c}
\toprule
Method & Accuracy (\%) & Gain (pp) \\
\midrule
W4 (Baseline) & 64.57 & - \\
+ Logit-Space Adapter & 64.57 & +0.00 \\
+ \textbf{Recti-Q} (Ours) & \textbf{65.74} & \textbf{+1.17} \\
\bottomrule
\end{tabular}
\end{table}

\subsubsection{Effect of Adapter Rank ($r$)}

Finally, we analyze the parameter-efficiency of Recti-Q by varying the LoRA rank $r$. We test this on the most challenging DeiT-s sketch domain, with results shown in Table \ref{tab:ablation_rank}.

The data reveals a clear ``sweet spot'' of diminishing returns. A minimal rank of $r=4$ (adding only 0.006 MB) recovers +0.77 pp, which is over 74\% of the total gain achieved by the 16x larger $r=64$ adapter. Increasing the rank to $r=16$ (a mere 0.025 MB) captures nearly 90\% of the maximum recovery (+0.95 pp).

This is a critical finding: it confirms that our method is not just parameter-efficient, but extremely parameter-efficient. At merely 6 KB (0.006 MB), this adapter payload can be transmitted near-instantaneously over low-bandwidth cellular or satellite networks, serving as concrete proof that Recti-Q provides a highly practical, near-zero-cost solution for Over-The-Air (OTA) robustness patching in remote robotic fleets.

\begin{table}[h!]
\centering
\small
\caption{\textbf{Ablation: Effect of LoRA rank ($r$) on robustness.}
We analyze the accuracy vs. parameter cost on the worst-case DeiT-s sketch domain. The baseline W4 accuracy is 64.57\%.
}
\label{tab:ablation_rank}
\begin{tabular}{c c c c}
\toprule
Rank ($r$) & Adapter Size (MB) & Accuracy (\%) & Gain (pp) \\
\midrule
4 & 0.006 & 65.34 & +0.77 \\
8 & 0.012 & 65.41 & +0.84 \\
16 & 0.025 & 65.52 & +0.95 \\
32 & 0.050 & 65.59 & +1.02 \\
64 & 0.100 & 65.61 & +1.04 \\
\bottomrule
\end{tabular}
\end{table}

\section{Conclusion} 
\label{sec:conclusion}

We addressed a critical, yet under-explored, trade-off between model efficiency and deployment robustness. We showed that standard Post-Training Quantization (PTQ), a vital tool for efficiency on edge platforms, can substantially degrade out-of-distribution (OOD) robustness, creating a ``Quantization-Induced Robustness Gap.'' We provided evidence for this gap on both common corruptions (ImageNet-C) and domain shifts (PACS), demonstrating that models that appear robust on in-distribution data can fail silently and severely under shift.

To address this, we introduced \textbf{Recti-Q}, a framework that decouples model compression from robustness recovery. Instead of expensive re-training, Recti-Q freezes the efficient, quantized backbone and trains a single, lightweight robustness-repairing adapter ($<1\%$ overhead) using only source-domain data. Our approach is motivated by the hypothesis that PTQ perturbs the model's feature space, which can be corrected by a small, parameter-efficient adapter at the classifier head.

Our experiments validate this hypothesis. As shown in Table~\ref{tab:main_results_pacs} and Table~\ref{tab:main_results_imagenetc}, Recti-Q consistently recovers a significant portion of the robustness gap and, in some cases, even exceeds the original FP32 robustness, while retaining over 99\% of the PTQ model's memory savings and incurring negligible additional inference cost. 

The implications of this work are practical. Recti-Q challenges the assumption that robust quantization necessarily requires computationally expensive Quantization-Aware Training. Instead, one can take an ``off-the-shelf'' quantized model and apply a rapid, low-cost robustness patch, enabling low-bandwidth Over-The-Air (OTA) updates for deployment. 

\noindent\textbf{Limitations and Scope.} Recti-Q corrects quantization-induced feature distortion but cannot reconstruct information lost by the frozen backbone. Because the adapter acts only on the backbone's output features, its effectiveness depends on those features retaining usable class structure; under sufficiently aggressive quantization, this assumption may fail and head-only rectification may become ineffective. Our evaluation is limited to closed-set covariate and domain/style shifts (ImageNet-C and PACS); Recti-Q will be evaluated on semantic or open-set OOD and is not yet expected to recognize categories outside the original training taxonomy. 

\noindent
{\bf Acknowledgement: } This project is supported in part by Dr. Barry Mersky and Capital One E-Nnovate Endowed Professorships and UMD-ARL Cooperative Agreement.

{
    \small
    \bibliographystyle{IEEEtran}
    \bibliography{main}
}


\end{document}